\documentclass{article}
\pdfoutput=1
\usepackage{graphicx}
\usepackage[a4paper]{geometry}

\begin{document}
\date{}
\title{Variational Grid Setting Network}
\author{
Chuang, Yu-Neng\\
Department of Mathematical Sciences, National Chengchi University\\
Taipei, Taiwan\\
\texttt{n08041134@gmail.com}
\and
Huang, Zi-Yu\\
Institute of Industrial Engineering, National Taiwan University\\
Taipei, Taiwan\\
\texttt{moon.starsky37@gmail.com}
\and
Tsai, Yen-Lung\\
Deptartment of Mathematical Sciences, National Chengchi University\\
Taipei, Taiwan\\
\texttt{yenlung@nccu.edu.tw}
}
\maketitle

\begin{abstract}
We propose a new neural network architecture for automatic generation of missing characters in a Chinese font set. We call the neural network architecture the Variational Grid Setting Network which is based on the variational autoencoder (VAE) with some tweaks. The neural network model is able to generate missing characters relatively large in size ($256 \times 256$ pixels). Moreover, we show that one can use very few samples for training data set, and get a satisfied result.\\\\

\end{abstract}


\section{Introduction}
Chinese characters are abstractly aesthetic glyphs and usually considered as the most complex ones among all the other language characters. Creating Chinese fonts is a time-consuming task. We propose a new neural network architecture, the \emph{Variational Grid Setting Network} (VGSN), to generate missing Chinese characters from a Chinese font set. 

Our model is based on the variational autoencoder (VAE). The key differences between our model and the original VAE are we add a special layer which we called the \emph{grid setting layer} in the end, to generate large image ($256 \times 256$ pixels) of Chinese characters. Moreover, we modified the loss function to gain the best results.

The scenario is as follows: Font A has complete Chinese characters, at least the one we want, and Font B has some missing characters which we would like to get automatically from Font A. We pair some characters that are available in both Font A and Font B as the training dataset. After that, we put the one which is missing in font B but available in Font A into our network, and (hopefully) get the character in the style of font B.

Somewhat surprising observation is that we just need very few samples to generate a character image we desire. In our experiments, usually $5$ examples will be sufficient and we can get a high quality image. As a result, our model runs relatively fast so it has potential to help professional typographers to design new Chinese fonts.

\section{Architecture}
This model mainly contains three parts. The first part of our architecture is the encoder which task is to get the feature of Asian characters, that is,  given the image of character with dimension (h, w), find the latent vector of there feature. We use neural network with input layer format as (batch size, height of image, width of image, channel of image), and output layer of the size two, (mean, variation). This part uses $6$ convolution layer with the ReLU as activation function, $6$ batch normalization layer, a fully connect layer to generating basis, and two separate fully connect layer to calculate the latent basis and get the latent vector.\\

The second part is our grid setting layer which contains dense layer and reshape layer. This part will split the original image into separate partition. The dense layer  use to amplify the latent vector for each partition, and the reshape layer use to setting the grid line format.\\

The final part is the decoder which need to generating the image of each partition and combiming them. 
This part uses $5$ convolution transpose layer with the ReLU as activation function and same padding, $5$ batch normalization layer to generating the image. And  final output layer use a convolution transpose layer with only $1$ kernel to compile partitions together.

\begin{figure}[!h]
	\includegraphics[width=3in]{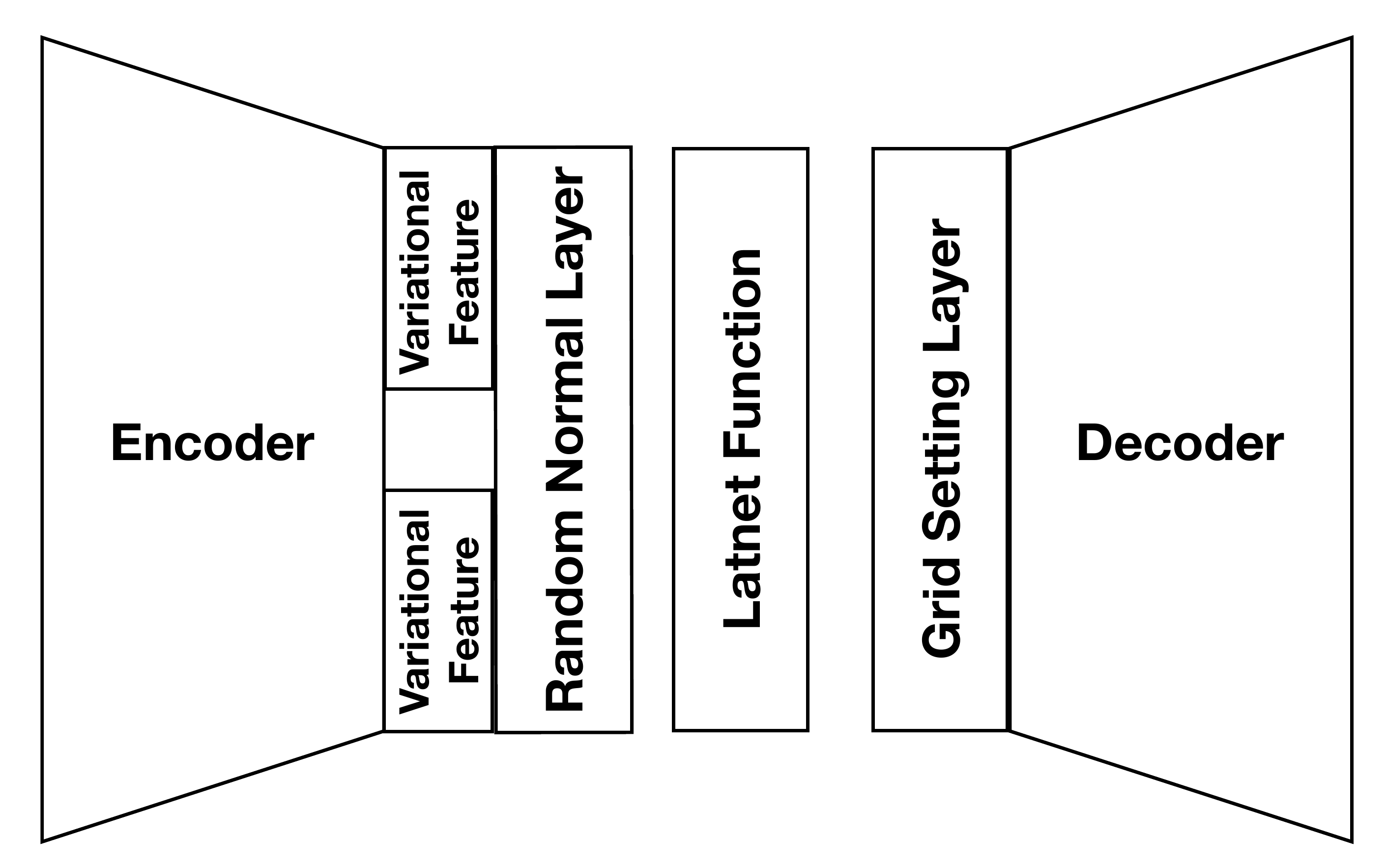}
	\centering
	\caption{Structure of Variational Grid Setting Network.}
\end{figure}

\subsection{Latent basis calculation}
As above, when we estimate the latent vector, we need to calculating the latent basis, respectively. Since the first part of our neural network reduction high dimension of original image, the basis value become critical when we need to decode these digit. In order to preserve the continuous property of basis, we use Gaussian normal distribution to get the feature of encoder. Then, we times the result of probability distribution with exponential of power of encoder variation and plus a mean of encoder which is more effective than the latent basis of  conventional VAE. That is, \\
$$\mu+\exp^{\sigma}\times\frac{1}{{ \sqrt {2\pi } }}e^{{{ - {x} ^2 } \mathord{\left/ {\vphantom {{ - 2\sigma ^2 } }} \right. \kern-\nulldelimiterspace} {2 }}}$$

where $\mu$ is the mean of encoder, and the $\sigma$ is the variational of encoder.\\

\subsection{Traing algorithm}
For training, the encoder use standard back-propagation algorithm.  Subsampling of these feature is modified the traditional convolution neural network architecture from [2]. Since Asian characters is continuous in lots of region, adding some strides and changing number of kernel is used to preserve the properties and reduce the parameter. To accelerating network training in normalization, our normalizing layer  based on [3] which reducing internal covariate shift.

After calculating latent basis, two dimension vector, called latent vector and denoted by $[\sigma ,\mu]$, generates new probability distribution. And the normalization is ensured by the sigmoid. Then we set the grid in our model,and we use 4$\times$ 4 and 8$\times$8 as experiment. We add some dense layer to preserve of probability distribution.

The decoder is based on the deconvolution neural network [4], and we add strides as same in encoder for symmetry. As well the same number of batch normalization layer.

When we train the model, our batch size cannot choose too large (we choose $32$ in most cases ), since the range of result need to same as domain. In order to avoid over-fitting, we need to shuffle the input data in each epoch.\\

\section{Experiment}
We use some open source data (like Source-Han-Sans and Source-Han-Serif) for our learning model. In our experiment, we generate those character in high dimensional graph and transfer the font style if we choose another one as feature. Two modes of grid setting are used, one is 4$\times$ 4 grid and another is 8$\times$ 8 grid, which means a 256$\times$256 image are split up into $16$ and $64$ parts where the latent vectors are trained respectively.

Given the learned models and mode, we propose three types of optimization methods: Stochastic Gradient Descend (SGD), Adaptive Moment Estimation (Adam) and RMSprop. To make sure we reduce the training time, we also test the traditional VAE, and we can discover that VGSN is faster than traditional one while generating same image in one epoch. 

Table 1 and Table 2 show that VGSN trains much faster than the original VAE in same condition. Therefore, it shows that VGSN can handle the fast generating process and avoid overfitting simultaneously.  And in Figure 2, we campare with VAE model which shows that VGSN is better than original VAE with grid setting layer. We use original VAE without grid setting layer at first, the outcome were all noise and overfitting. Therefore, we decide to use original VAE model to compare with VGSN. \\\\

\begin{center}
\begin{tabular}{cccc}
\hline
&SGD & Adam & RMSprop \\ \hline
4$ \times $4 grid  &87 sec. & 96 sec. & 95 sec. \\
8$ \times $8 grid &179 sec. & 177 sec. & 175 sec. \\
\hline
\end{tabular}
\end{center}
~~~~~~~~~~~~~~~~~~~~~~~~~~~~~Table 1: Time in original Variational Autoencoder\\\\\\

\begin{center}
\begin{tabular}{cccc}
\hline
&SGD & Adam & RMSprop \\ \hline
4$ \times $4 grid  &78 sec. & 77 sec. & 74 sec. \\
8$ \times $8 grid  &148 sec.  & 157 sec. & 154 sec. \\
\hline
\end{tabular}
\end{center}
~~~~~~~~~~~~~~~~~~~~~~~~~~~~~~~~~~~~~~~~~~~~~~~Table 2: Time in VGSN\\\\\\

In terms of different model and optimizer for the training model, mean squared error is choosed as loss function to measure the loss in all combination of them. The results of losses plot as follows (Figure 3 and Figure 4). According to the following results, if we use Adam as our optimizer, the learning process will avoid being overfitting. In the case of SGD, the model learns slower because of early convergence. And if  RMSprop is given as our optimizer, the model learns well at the early epoch of training process, but fail at the late epoch.
\begin{figure}
	\begin{center}
		(a). Traditional VAE (left) and VGSN (right) under 4$ \times $4 grid. \\
		
		\includegraphics[width=1in]{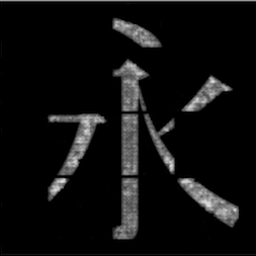}
		\includegraphics[width=1in]{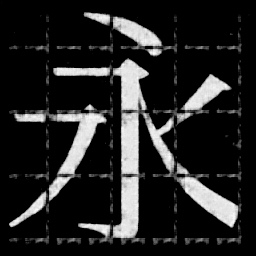}
		
		(b). Traditional VAE (left) and VGSN (right) under 8$ \times $8 grid. \\
		
		\includegraphics[width=1in]{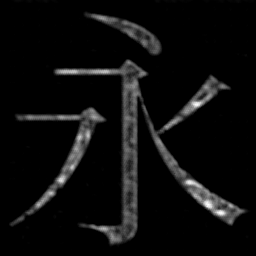}
		\includegraphics[width=1in]{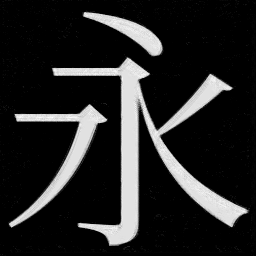}
	\end{center}
	\caption{Compare the figure generate in VGSN (the right row) wih the figure generate in original VAE  model (in the left row) in same condition and under same grid setting. The pronunciation of the characters are all yong.}
\end{figure}

\begin{figure}[!h]
	\includegraphics[width=3in]{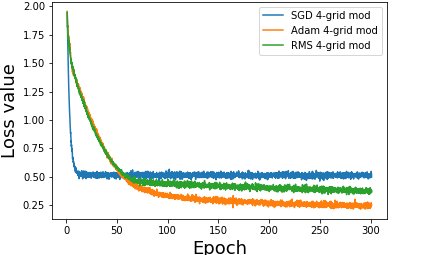}
	\centering
	\caption{Loss in 4$\times$4 grid with different optimizer.}

	\includegraphics[width=3in]{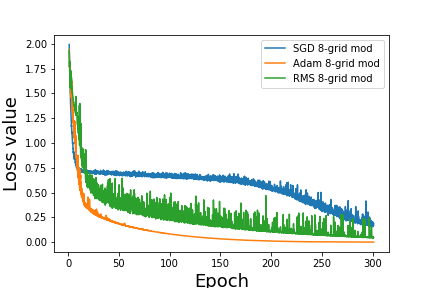}
	
	\centering
	\caption{Loss in 8$\times$8 grid with different optimizer.}
\end{figure}

\subsection{Optimizer selection}
In this work, we choose Source-Han-Sans and Source-Han-Serif as our data set within given three different optimizers and try to figure out which one takes the advantage of generating image.  

First, given SGD as our optimizer, we realize that the radical part is certainly well-learning, but few of Chinese characters are somewhere fuzzy. And then given Adam as our optimizer, the main structure is clear, and side area are less fuzzy than SGD.  Last, given RMSprop as our optimizer. Although main structure is clear and recognizable, some grid edges are not well-connected and some detail are missing.

In conclusion, no matter which grid we set, Adam plays an essential role. Viewing those outcome(Figure 4 and Figure 5), Adam might not only be the fastest one, but it makes our model to learn more efficient than other two optimizers. 
\\\\

\subsection{Grid setting selection}
In order to figure out which grid setting performs well, we choose Source-Han-Sans and Source-Han-Serif as our data set within fixing the optimizers. 

First, fixed the optimizer, although the main part in both grids are nearly well-learning, it seems that 4$\times$4 grid still has grid segment line, but 8$\times$8 grid did well on this part. Second, we can observe that 8$\times$8 grid solve the connected problem better than 4$\times$4 grid.

In conclusion, no matter which optimizer we used, 4$\times$4 grid is fuzzy at the edge area,and some details are lost at the connected side of each grid.\\

\begin{figure}[!h]
\centering
(a). Given SGD as optimizer under four-times-four grid.

\centering
\includegraphics[width=1.1in]{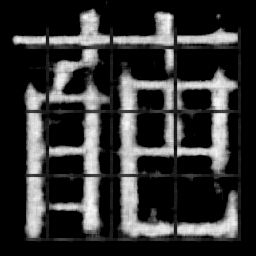}
\includegraphics[width=1.1in]{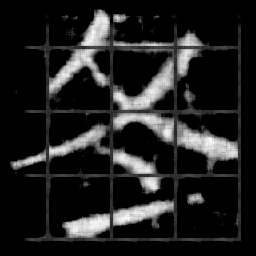}
\includegraphics[width=1.1in]{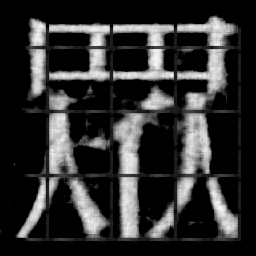}

(b). Given RMSprop as optimizer under four-times-four grid.\\

\includegraphics[width=1.1in]{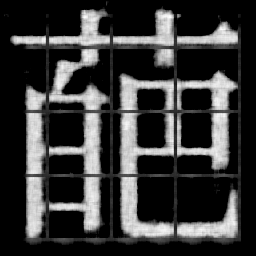}
\includegraphics[width=1.1in]{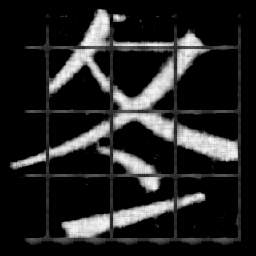}
\includegraphics[width=1.1in]{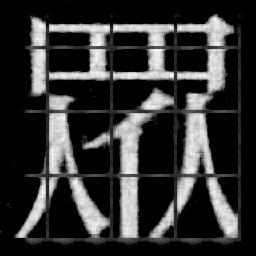}

(c). Given Adam as optimizer under four-times-four grid.\\

\includegraphics[width=1.1in]{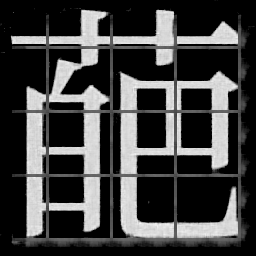}
\includegraphics[width=1.1in]{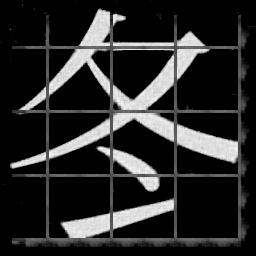}
\includegraphics[width=1.1in]{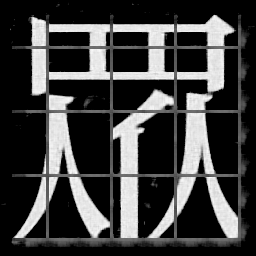}

\caption{Outcome of Chinese character generated by using SGD, Adam and RMSprop respectively as the optimizer under four-times-four grid. The pronunciation of these characters from left to right are pa, dong and zhong.}
\end{figure}

\begin{figure}[!h]
\centering
(a). Given RMSprop as optimizer under eight-times-eight grid.\\

\centering
\includegraphics[width=1.1in]{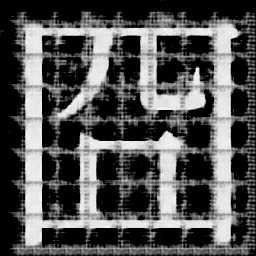}
\includegraphics[width=1.1in]{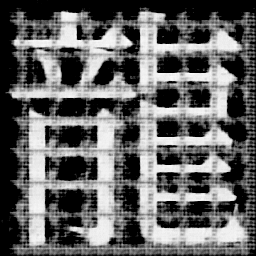}
\includegraphics[width=1.1in]{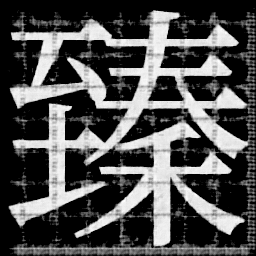}

(b).  Given SGD as optimizer under eight-times-eight\\
\includegraphics[width=1.1in]{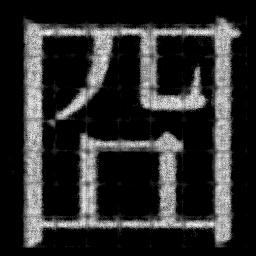}
\includegraphics[width=1.1in]{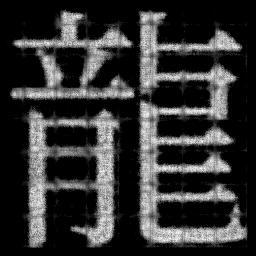}
\includegraphics[width=1.1in]{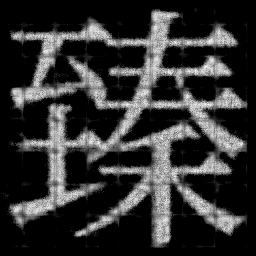}

(c).  Given Adam as optimizer under eight-times-eight grid.\\
\centering
\includegraphics[width=1.1in]{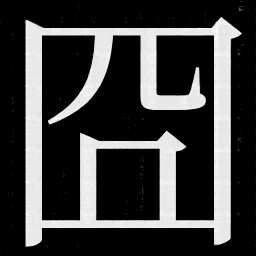}
\includegraphics[width=1.1in]{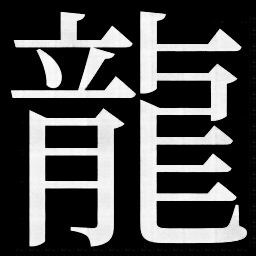}
\includegraphics[width=1.1in]{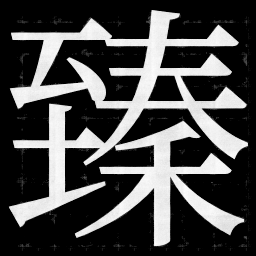}

\centering
\caption{Outcome of Chinese character generated by using SGD, Adam and RMSprop respectively as the optimizer under eight-times-eight grid. The pronunciation of these characters from left to right are jiong, long and zhen.}
\end{figure}

\subsection{Font transfer}
In this work, we try to generate a style-transferred Chinese character by using two different font as our data set. Adam is selected as our optimizer and 8$\times$8 grid is chosen as our grid setting parameter. The results show in Figure 6 are transferred by taking Source-Han-Sans-J being our training data and using Source-Han-Serif-K being our testing set. 

First, we train our model to learn the feature of $4$ characters of Source-Han-Sans-Japan, and then we apply the feature to generate the characters of Source-Han-Serif-Korean. Moreover, we also try to minimize our training data set and generate the Chinese characters those are not in the training set. Figure 6 shows the style transfer outcome of VGSN. In part a, we pick up the first four characters and generate the last character by training under our model. 

In order to highlight our training data is not overlapping our testing data, Figure 6 also shows that we use two different fonts and choose the data randomly from our training font. Training data is announced as follows. The characters are automatically generated by VGSN and not the samples in original training data.

The pronunciation of following example characters in first three row of part a are qian, shan, fei and jue from left to right. In forth row of part a is niao. And in first three row of part b are yong, qin, zhi, she and he from left to right. In other rows from top to bottom are pang, yi, tao and kan.  

\begin{figure}[!tp]
(a)
\begin{center}
\includegraphics[width=2.5in]{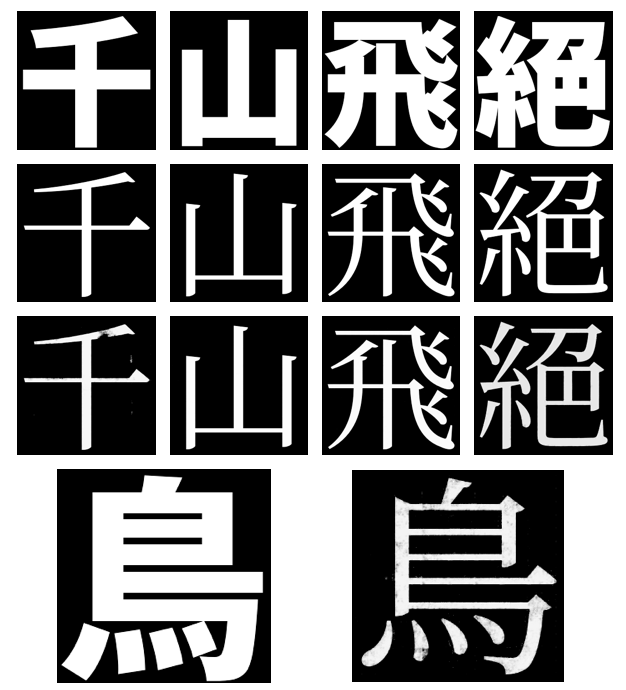}
\end{center}

In above figure, the first row is input, which uses Source-Han-San. The second row is ground truth corresponding to the input, which uses Source-Han-Serif. The third row is the results we generated by VSGN. After training, the left hand side in the last row is the character we want to predict, and the right hand side is the result we generated by our model.\\

(b)
\begin{center}
\includegraphics[width=2.85in]{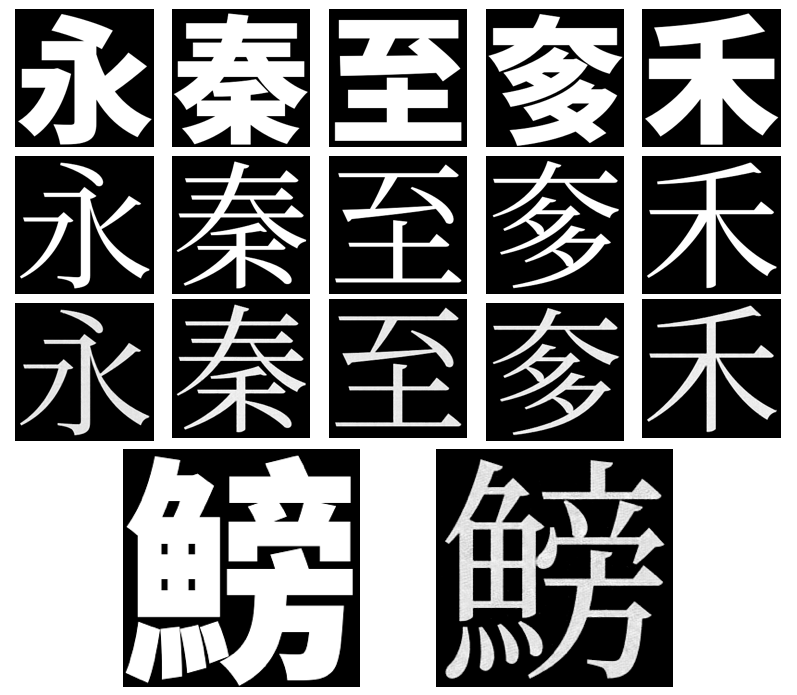}
\includegraphics[width=2.05in]{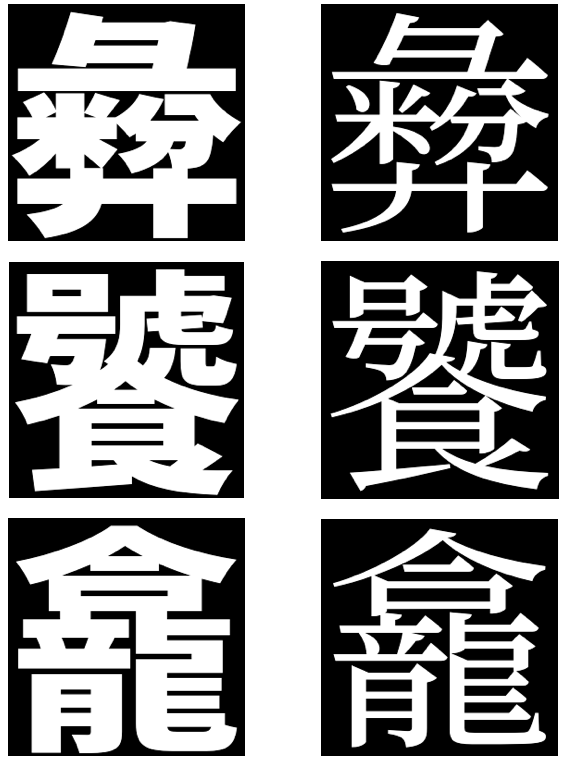}
\end{center}

In this figure, we use those characters which contains more strokes in regular script as input. Then, we can generate more complicated by VSGN.
\caption{Visualization of the samples and outcome from automatic generating the Chinese character.}
\end{figure} 

\pagebreak
\section{Conclusion}
We introduce Variational Grid Setting Network (VGSN) in this paper. Our approach is that we can not only generate a high dimensional image within 256$\times$256 pixel through grid setting layer, but also can perform the style transfer between two different fonts by using very few samples.

According to the result of our experiment, we discover that if we choose Adam as our optimizers and 8$\times$8 grid as our grid setting, the outcome is more clear than the other 8 kinds of combination. 

Furthermore, it is possible to use our model to deal with word missing problem. Due to the result of font transfer, we can generate the words which do not exist in the new font. Word selection is available and few training data is required in automatically generating the characters. Therefore, we can produce the high quality image of specific characters  of rebuilding the missing characters. It surely benefits the typographers to reduce the time while designing new Chinese fonts in the future.

\end{document}